\documentclass{article}

\usepackage[preprint]{neurips_2026}

\usepackage[utf8]{inputenc} 
\usepackage[T1]{fontenc}    
\usepackage{hyperref}       
\usepackage{url}            
\usepackage{booktabs}       
\usepackage{amsfonts}       
\usepackage{nicefrac}       
\usepackage{microtype}      
\usepackage{xcolor}         
\usepackage{amsmath}
\usepackage{amssymb}
\usepackage{bm}
\usepackage{wrapfig}
\usepackage{graphicx}
\usepackage{threeparttable}

\usepackage{amsthm}
\usepackage{algorithm}
\usepackage{algorithmic}
\usepackage[capitalize,nameinlink]{cleveref}
\usepackage{enumitem}
\usepackage{makecell}
\usepackage{multirow}
\usepackage{tikz}

\newcolumntype{Y}{>{\centering\arraybackslash}X}

\crefname{equation}{Eq.}{Eqs.}
\Crefname{equation}{Eq.}{Eqs.}
\crefname{figure}{Fig.}{Figs.}
\Crefname{figure}{Fig.}{Figs.}
\crefname{table}{Tab.}{Tabs.}
\Crefname{table}{Tab.}{Tabs.}
\crefname{section}{Sec.}{Secs.}
\Crefname{section}{Sec.}{Secs.}
\crefname{appendix}{App.}{Apps.}
\Crefname{appendix}{App.}{Apps.}

\newcommand\app[0] {App.}
\newcommand\ie[0] {\emph{i.e., }}
\newcommand\eg[0] {\emph{e.g., }}

\definecolor{color_red}{RGB}{255,0,0}
\definecolor{color_green}{RGB}{0,255,0}
\definecolor{color_blue}{RGB}{0,0,255}
\definecolor{color_gray}{RGB}{127,127,127}

\title{Tensor-Train Joint Modeling for Few-Step Masked Diffusion}

\author{%
  Byoungkwon Kim \\
  KAIST \\
  \texttt{ssamt@kaist.ac.kr} \\
  \And
  Minhyuk Sung \\
  KAIST \\
  \texttt{mhsung@kaist.ac.kr} \\
}

\begin{document}

\maketitle

\begin{abstract}
Discrete diffusion promises orders-of-magnitude faster generation than autoregressive (AR) models for sequential discrete data, yet its full potential of few-step generation has remained out of reach due to a fundamental structural limitation. The conditional-independence assumption underlying current discrete diffusion models introduces a systematic \emph{parallelization bias} that compounds with the number of tokens unmasked per step, becoming severe in the few-step regime that fast generation requires. We address this with the first framework for explicit joint distribution modeling in discrete diffusion via \emph{tensor decomposition}, which represents the conditional clean distribution as a low-rank tensor with controllable expressivity. The framework supports both Canonical Polyadic (CPD) and Tensor-Train (TTD) decompositions, and we identify a structural bias of TTD toward dependencies between nearby tokens, formalized through Oseledets' theorem relating TT-rank to unfolding-matrix rank, which is well-suited to sequential data such as natural language and line notations for molecular data. To enable efficient generation, we present an iterative \emph{marginal inference} procedure with specialization for predetermined position schedules. Our framework integrates into pretrained MDMs through lightweight fine-tuning, yielding substantial improvements in few-step generation at a fraction of the cost of training from scratch.
\end{abstract}
\section{Introduction}

Discrete diffusion has emerged as a compelling alternative to AR models for generating sequential discrete data such as text and proteins~\citep{austin2021d3pm, lou2024sedd, sahoo2024mdlm, shi2024md4}. Its key promise is parallel token generation, producing multiple tokens per step rather than one. Most notably, Masked Diffusion Models (MDMs)~\citep{sahoo2024mdlm, shi2024md4} have scaled to large language models~\citep{nie2025llada, ye2025dream} with competitive quality and faster generation.

In practice, however, this speed potential remains unrealized, as pushing toward few-step generation systematically degrades sample quality. The root cause is structural; to remain tractable, discrete diffusion models represent the conditional clean distribution $p_\theta(\bm{x} | \bm{x}_t)$ under the conditional-independence assumption $p_\theta(\bm{x} | \bm{x}_t) = \prod_{i=1}^{N} p_\theta(x^i | \bm{x}_t)$. This assumption introduces a systematic \emph{parallelization bias}~\citep{zhang2026generationorderparalleldecoding} when multiple tokens are unmasked simultaneously. The bias compounds with the number of tokens unmasked per step, becoming especially severe in the few-step regime that fast generation requires.

Resolving this bias requires modeling the joint distribution $p_\theta(\bm{x} | \bm{x}_t)$ explicitly. However, na\"{i}ve joint modeling is infeasible since the joint over $N$ tokens with vocabulary size $V$ forms an $N$-dimensional tensor with $V^N$ entries. Some existing approaches therefore sidestep direct joint modeling, introducing latent factors that induce token correlations~\citep{hayakawa2025di4c, xie2026vadd} or borrowing dependence from auxiliary autoregressive models~\citep{liu2025dcd, xu2025edlm}. Compared to AR-auxiliary approaches, our method models the joint distribution explicitly as a tractable parametric family within the discrete diffusion model itself, without requiring any external pretrained model. Also, compared to latent-variable approaches, our explicit parameterization is orthogonal to their implicit one, and the two can be combined for further gains. Concurrent work~\citep{li2026breakingfactorizationbarrierdiffusion} addresses the same factorization barrier by augmenting the factorized neural output with a tractable probabilistic circuit prior, requiring manual structural design for sequential locality, whereas Tensor-Train decomposition provides the same tractability with a theoretically grounded locality bias.

In this work, we introduce a framework that explicitly parameterizes the joint distribution $p_\theta(\bm{x} | \bm{x}_t)$ via \emph{tensor decomposition}~\citep{kolda2009tensor}. Compared to prior work, our framework provides explicit joint modeling within the diffusion model itself, without relying on auxiliary models. Our contributions are four-fold.
\begin{itemize}[leftmargin=*, itemsep=2pt, topsep=2pt]
    \item \textbf{First explicit joint modeling via tensor decomposition.} To our knowledge, this is the first work to parameterize the conditional clean distribution of discrete diffusion via tensor decomposition, strictly generalizing standard MDMs as the rank-one case.
    \item \textbf{TTD for sequential structure.} We develop the framework for both Canonical Polyadic Decomposition (CPD)~\citep{hitchcock1927cpd} and Tensor-Train Decomposition (TTD)~\citep{oseledets2011ttd}, identifying a structural bias of TTD toward dependencies between nearby tokens, formalized through Oseledets' theorem relating TT-rank to unfolding-matrix rank.
    \item \textbf{Efficient sampling via marginal inference.} Direct sampling from a joint of size $V^N$ is intractable. We present a chain-rule procedure built on iterative \emph{marginal inference} via caching and parallel prefix sums, with specialization for predetermined position schedules.
    \item \textbf{Lightweight fine-tuning of pretrained MDMs.} Our framework integrates into pretrained MDMs through small architectural modifications initialized to preserve the marginal predictions, yielding substantial few-step improvements at a fraction of the cost of training from scratch.
\end{itemize}

In experiments on text generation (OpenWebText~\citep{gokaslan2019owt}, LM1B~\citep{chelba2014lm1b}) and molecule generation (QM9~\citep{blum2009qm9_1,rupp2012qm9_2}), our framework delivers substantial few-step improvements across multiple base models, including MDLM~\citep{sahoo2024mdlm} and VADD~\citep{xie2026vadd}. As predicted by our theory, TTD consistently outperforms CPD on these sequential domains, with the largest gains concentrated in the few-step regime. For instance, TTD-based fine-tuning of VADD reduces 8-step generative perplexity on OpenWebText by 32.7\% over vanilla VADD, while being only 1.7\% slower at 128 timesteps.

\section{Related Work}
Several lines of work attempt to address the parallelization bias arising from the conditional-independence assumption in discrete diffusion. The central challenge is the exponential size of the joint distribution: an $N$-token sequence over vocabulary $V$ admits $V^N$ possible configurations, making direct modeling intractable.

Auxiliary-model approaches distill dependency information from separately trained autoregressive models via copula-based reweighting~\citep{liu2025dcd} or energy-based reweighting~\citep{xu2025edlm}. While effective, these methods inherit the inference cost of an external model and become inapplicable in domains lacking strong AR models, such as generation of line notations for molecular data. Our framework, in contrast, requires no additional model.

Latent-variable approaches introduce latent factors that induce token correlations through marginalization, either via a continuous mixing variable~\citep{hayakawa2025di4c} or a VAE-style latent vector~\citep{xie2026vadd}; our explicit modeling is orthogonal to these implicit approaches and can be combined with them for further gains.

Sampling strategies mitigate the bias without modifying the model, selecting positions via low-discrepancy schedules~\citep{besnier2025halton} or adaptively bounding the number of simultaneously unmasked tokens through entropy criteria~\citep{benhamu2025ebsampler}; these strategies integrate naturally within our framework.

Concurrent work~\citep{li2026breakingfactorizationbarrierdiffusion} couples the factorized neural output with a tractable probabilistic circuit prior, addressing the same factorization barrier with a different parameterization. Although probabilistic circuits offer general-purpose tractable inference, they require manual structural design for sequential locality, whereas Tensor-Train decomposition provides the same tractability with a theoretically grounded locality bias (\cref{sec:ttd}), making it particularly suited to natural language and line notations for molecular data without further structural engineering.

The parallelization bias has also been studied in the context of speculative decoding for \emph{autoregressive} language models, where \citet{basharin2025fasterlanguagemodelsbetter} were the first to apply Canonical Polyadic Decomposition for explicit token-wise dependency modeling. We extend this idea to discrete diffusion for the first time, with two main technical contributions beyond their work: (1) identifying Tensor-Train Decomposition as particularly suited to sequential structure, with a theoretically grounded inductive bias toward dependencies between nearby tokens; and (2) presenting an efficient sampling procedure that introduces minimal overhead when sampling from the joint distribution.

\section{Background \& Overview}
\subsection{Background: Masked Diffusion Models (MDM)}

A Masked Diffusion Model (MDM)~\citep{sahoo2024mdlm} is a discrete diffusion model whose forward process is an absorbing process. The absorbing token is commonly referred to as the mask token $m$ and thus lends the model its name. The forward process maps the data distribution $p(\bm{x})$ to the prior distribution, which is a Dirac delta concentrated at the fully masked sequence $\mathbf{m} = [m, m, \ldots, m]$, as $t$ progresses from $0$ to $1$. Formally,
\begin{equation}
    q(x_t^i | \bm{x}) = t\,\delta_{m}(x_t^i) + (1-t)\,\delta_{x^i}(x_t^i),
\end{equation}
where $x_t^i \in [V] = \{1, 2, \ldots, V\}$ denotes the $i$-th token at time $t \in [0,1]$, $\bm{x} = [x^1, x^2, \ldots, x^N]$ is the clean sequence, $\bm{x}_t = [x_t^1, \ldots, x_t^N]$ is its partially masked version, and $\delta_a$ denotes the point mass at $a$.

The model is trained to predict the original values of the masked tokens in a partially masked sequence, typically under an assumption of conditional independence across tokens. Namely,
\begin{equation}
    p_\theta(\bm{x} | \bm{x}_t) = \prod_{i=1}^{N} p_\theta(x^i | \bm{x}_t).
    \label{eq:mdm-factorization}
\end{equation}
Given a partially masked sequence $\bm{x}_t$, for every position $i$ and every $x^i \in [V]$, the model predicts the marginal probability $p_\theta(x^i | \bm{x}_t)$ that the $i$-th token of the clean sequence equals $x^i$. Under the factorization, the negative log-likelihood (NLL) loss
\begin{equation}
    \mathcal{L}_{\text{NLL}}
      = \mathbb{E}_{t \sim U(0,1),\; \bm{x} \sim p(\bm{x}),\; \bm{x}_t \sim q(\bm{x}_t | \bm{x})}
            \!\left[-\log p_\theta(\bm{x} | \bm{x}_t)\right]
    \label{eq:nll}
\end{equation}
reduces to the sum of per-position marginal NLLs:
\begin{equation}
    \mathcal{L}_{\text{marginal}}
      = \mathbb{E}\!\left[\sum_{i=1}^{N} -\log p_\theta(x^i | \bm{x}_t)\right].
\end{equation}

Generation proceeds by initializing a fully masked sequence and iteratively unmasking tokens until a clean, mask-free sequence is obtained.

\subsection{Method Overview}
Despite the promising progress of masked diffusion models, a key limitation lies in the conditional-independence assumption (\cref{eq:mdm-factorization}) imposed on $p_\theta(\bm{x} | \bm{x}_t)$, which introduces a systematic \emph{parallelization bias} \citep{zhang2026generationorderparalleldecoding} that becomes especially severe in the few-step regime. Explicitly modeling the joint distribution $p_\theta(\bm{x} | \bm{x}_t)$, however, is fundamentally a problem of scale. In continuous diffusion, the joint admits a tractable parameterization as a multivariate Gaussian. In the discrete setting, by contrast, the joint corresponds to an $N$-dimensional tensor of size $V^N$, which rapidly becomes intractable as the vocabulary size $V$ and the sequence length $N$ grow.

To address this challenge, we propose to model $p_\theta(\bm{x} | \bm{x}_t)$ explicitly via \emph{tensor decomposition}. In~\cref{sec:method}, we investigate two decompositions particularly suited to this purpose---Canonical Polyadic Decomposition (CPD)~\citep{hitchcock1927cpd} and Tensor-Train Decomposition (TTD)~\citep{oseledets2011ttd}---both supporting efficient training and sampling with only $O(VN)$ memory. We describe how each parameterizes the joint distribution and how generation is carried out under it. Our analysis identifies a structural limitation of CPD that motivates the use of TTD, whose form is naturally aligned with dependencies between \emph{nearby} tokens, a property especially well suited to sequential data such as natural language \citep{ebeling1994longrangecorrelations} and line notations for molecular data. Note that either decomposition reduces to the standard marginal probability formulation (\cref{eq:mdm-factorization}) in the rank-one special case.
To our knowledge, this is the first work introducing tensor decomposition to joint probability modeling in discrete diffusion.

\section{Joint Probability Modeling via Tensor Decomposition}
\label{sec:method}

In this section, we describe how the joint distribution $p_\theta(\bm{x} | \bm{x}_t)$ can be modeled using two tensor decompositions: Canonical Polyadic Decomposition (CPD; \cref{sec:cpd}) and Tensor-Train Decomposition (TTD; \cref{sec:ttd}). We discuss the advantages of TTD over CPD in capturing dependencies between nearby tokens, a property crucial for many practical domains such as text and line notations for molecular data. Building on the formulations, \cref{sec:sampling} describes the sampling procedure under either decomposition.

\begin{figure}
  \centering
  \includegraphics[width=1\linewidth]{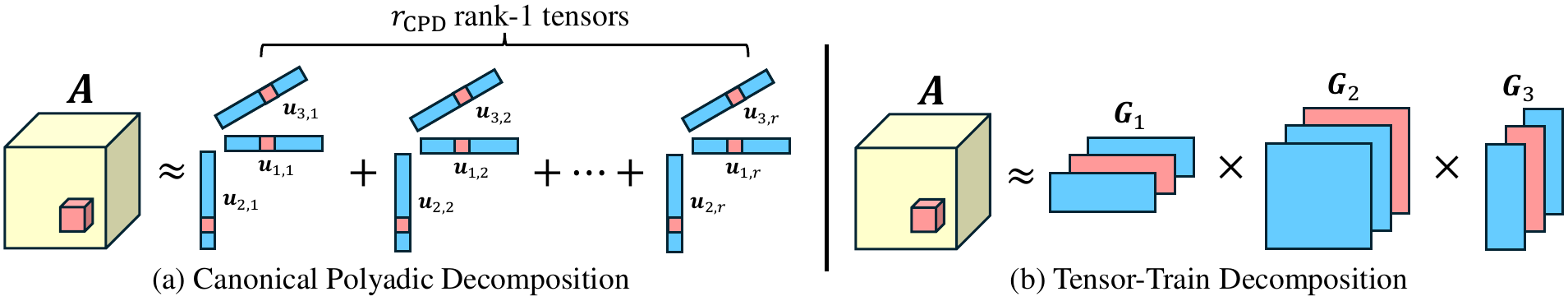}
  \caption{Comparison of tensor decomposition methods: Canonical Polyadic Decomposition (CPD) and Tensor Train Decomposition (TTD).}
  \label{fig:decompositions}
\end{figure}

\subsection{Modeling via Canonical Polyadic Decomposition (CPD)}
\label{sec:cpd}

Given an order-$N$ tensor $\bm{A}$, Canonical Polyadic Decomposition (CPD)~\citep{hitchcock1927cpd} approximates $\bm{A}$ by a sum of $r$ rank-1 tensors, each formed as the outer product of $N$ vectors:
\begin{equation}
\bm{A}(\nu_1, \ldots, \nu_N) \approx \sum_{\alpha=1}^{r} \prod_{i=1}^{N} \bm{u}_{i, \alpha}(\nu_i),
\end{equation}
where each $\bm{u}_{i, \alpha} \in \mathbb{R}^{V}$ is a one-dimensional factor vector, and $r$ is referred to as the \emph{CP-rank}, which we may write as $r_{\text{CPD}}$.
Building on this, we model the joint distribution $p_\theta(\bm{x} | \bm{x}_t)$ over a sequence of $N$ tokens via CPD, with an additional rank weight (as in the autoregressive setting of \citet{basharin2025fasterlanguagemodelsbetter}):
\begin{equation}
p_\theta(\bm{x} | \bm{x}_t) = \sum_{\alpha=1}^{r} w^{(\alpha)}_\theta(\bm{x}_t) \prod_{i=1}^{N} p^{(\alpha, i)}_\theta(x^i | \bm{x}_t),
\label{eq:cpd}
\end{equation}
where $w^{(\alpha)}_\theta(\bm{x}_t) \geq 0$ is the weight produced for the $\alpha$-th rank-1 component, and $p^{(\alpha, i)}_\theta(x^i | \bm{x}_t) \geq 0$ is the corresponding output for token $x^i$ at position $i$ in that component.
For normalization, we enforce
\begin{equation}
    \sum_{\alpha=1}^{r} w^{(\alpha)}_\theta(\bm{x}_t) = 1, \qquad \sum_{x^i=1}^{V} p^{(\alpha, i)}_\theta(x^i | \bm{x}_t) = 1,
\end{equation}
which together ensure that $p_\theta(\bm{x} | \bm{x}_t)$ is a valid probability distribution while maintaining expressibility (proof in Appendix~\ref{app:normalization_constraints}). These constraints further allow $p^{(\alpha, i)}_\theta(x^i | \bm{x}_t)$ to be interpreted as the marginal probability of $x^i$ at position $i$ within the $\alpha$-th component.

Note that when $r=1$, the formulation reduces to the standard MDM, which predicts only per-position marginal probabilities. Moreover, any lower-rank representation with $r' < r$ can be recovered by zeroing the surplus weights, so the family of joint distributions expressible at rank $r$ strictly contains those expressible at any lower rank.

In practice, $r$ must be kept small for computational and memory efficiency. To maximize the benefit of a low-rank approximation, the key is to model inter-token dependencies in a manner aligned with the structure of the data modality. For natural language, it is well known that \emph{nearby} tokens are the most strongly correlated \citep{ebeling1994longrangecorrelations}. To capture such local dependencies effectively, we next introduce Tensor-Train Decomposition (TTD) and discuss its advantages over CPD.

\subsection{Modeling via Tensor-Train Decomposition (TTD)}
\label{sec:ttd}

Tensor-Train Decomposition (TTD)~\citep{oseledets2011ttd} approximates an order-$N$ tensor $\bm{A}$ by a sequence of \emph{cores} $\bm{G}_1, \ldots, \bm{G}_N$, with each entry obtained as a product of the corresponding matrix slices:
\begin{equation}
\bm{A}(\nu_1, \ldots, \nu_N) \approx \bm{G}_1(\nu_1)\, \bm{G}_2(\nu_2) \cdots \bm{G}_N(\nu_N),
\end{equation}
where each core $\bm{G}_i \in \mathbb{R}^{V \times r_{i-1} \times r_i}$ has matrix slices $\bm{G}_i(\nu_i) := \bm{G}_i(\nu_i, :, :) \in \mathbb{R}^{r_{i-1} \times r_i}$, with boundary ranks $r_0 = r_N = 1$ so that the right-hand side is a scalar. The tuple of internal ranks $(r_1, r_2, \ldots, r_{N-1})$ is referred to as the \emph{TT-rank} of the decomposition.

We adopt this form to model the joint distribution $p_\theta(\bm{x} | \bm{x}_t)$, identifying each tensor mode with a sequence position. Each core $\bm{G}_{i,\theta}(x^i | \bm{x}_t) \in \mathbb{R}_{\geq 0}^{r_{i-1} \times r_i}$ is conditioned on the masked input. For practical implementation, we let every core share the common shape $V \times r \times r$, i.e., $r_0 = r_1 = \cdots = r_N = r$ (which we may also write as $r_{\text{TTD}}$), and contract the boundary dimensions with all-ones vectors $\bm{1}_r \in \mathbb{R}^{r}$ to recover a scalar:
\begin{equation}
p_\theta(\bm{x} | \bm{x}_t) = \tfrac{1}{r} \bm{1}_r^{\top}\, \bm{G}_{1,\theta}(x^1 | \bm{x}_t)\, \bm{G}_{2,\theta}(x^2 | \bm{x}_t) \cdots \bm{G}_{N,\theta}(x^N | \bm{x}_t)\, \bm{1}_r.
\label{eq:ttd}
\end{equation}
To ensure that $p_\theta(\bm{x} | \bm{x}_t)$ is a valid probability distribution, we impose the following normalization constraint on each core, which also maintains expressibility (proof in Appendix~\ref{app:normalization_constraints}):
\begin{equation}
\sum_{x^i, k} \bm{G}_{i,\theta}(x^i, j, k | \bm{x}_t) = 1, \quad \forall i, j.
\end{equation}
As with CPD, when $r=1$, the formulation reduces to the standard MDM, which predicts only per-position marginal probabilities. Moreover, any lower-rank representation with $r' < r$ can be recovered by zeroing surplus columns of the cores, so the family of joint distributions expressible at rank $r$ strictly contains those expressible at any lower rank.

A key advantage of TTD over CPD is its structural bias toward modeling dependencies between nearby tokens. To make this precise, we recall a result from \citet{oseledets2011ttd} that connects each TT-rank $r_i$ to the dependency structure of $\bm{A}$.

For each split position $i \in \{1, \ldots, N-1\}$, partition the $N$ token positions into two contiguous blocks: a \emph{left block} consisting of positions $1, \ldots, i$, and a \emph{right block} consisting of positions $i+1, \ldots, N$. The \emph{unfolding matrix} $\bm{A}_i$ at this split is obtained by reshaping $\bm{A}$ into a matrix whose rows are indexed by the joint values of the left block and whose columns by those of the right:
\begin{equation}
\bm{A}_i \in \mathbb{R}^{V^i \times V^{N-i}}, \qquad \bm{A}_i\!\bigl([\nu_1, \ldots, \nu_i],\,[\nu_{i+1}, \ldots, \nu_N]\bigr) = \bm{A}(\nu_1, \ldots, \nu_N).
\end{equation}
The rank of $\bm{A}_i$ captures how much information must flow across the cut at position $i$: a small $\operatorname{rank}(\bm{A}_i)$ means that the joint values of the left block can be summarized into a low-dimensional vector that fully determines the distribution over the right block, indicating that the dependence between the two sides is limited.

\citet[Theorem~2.1]{oseledets2011ttd} establishes that the minimum TT-rank required at level $i$ is exactly the rank of the corresponding unfolding matrix:
\begin{equation}
r_i = \operatorname{rank}(\bm{A}_i).
\end{equation}
Intuitively, this is because any TT representation of $\bm{A}$ already factorizes $\bm{A}_i$ into a product of a $V^i \times r_i$ matrix (built from the first $i$ cores) and a $r_i \times V^{N-i}$ matrix (built from the remaining cores), so $r_i \geq \operatorname{rank}(\bm{A}_i)$; the reverse direction is shown constructively in the same theorem.

The crucial implication is that each $r_i$ depends only on the dependence structure across the $i$-th cut, not on the overall complexity of $\bm{A}$. For sequential data dominated by \emph{local} dependencies, the unfolding ranks $\operatorname{rank}(\bm{A}_i)$ can remain small at every cut and not scale with the sequence length, so a small TT-rank $r$ can suffice for an accurate approximation. CPD, by contrast, treats all positions symmetrically and offers no analogous reduction when locality is the dominant structure of the data.

As a concrete illustration of how extreme this asymmetry can be, Appendix~\ref{app:example_of_ttd_advantage} constructs a specific distribution over $2N$ tokens for which $r_\text{CPD} = 2^N$ but $r_\text{TTD} = 2$.

\subsection{Efficient Sampling via Iterative Marginal Inference}
\label{sec:sampling}

The tensor-decomposed forms (CPD and TTD) of $p_\theta(\bm{x} | \bm{x}_t)$ admit efficient evaluation at any $\bm{x}$, so training proceeds identically to a standard MDM with the same NLL loss (\cref{eq:nll}). Sampling, however, is nontrivial: naively sampling from $p_\theta(\bm{x} | \bm{x}_t)$ would require evaluating it at all $V^N$ possible sequences, an exponential cost in the sequence length $N$. For efficient sampling, we present a general procedure based on the chain rule, followed by a special case that exploits additional structure.

The chain rule of probability decomposes the joint into a sequence of conditionals. For any ordering $(i_1, \ldots, i_K)$ of the $K$ positions,
\begin{equation}
p(x^{i_1}, \ldots, x^{i_K} | \bm{x}_t) = \prod_{j=1}^{K} p(x^{i_j} | x^{i_1}, \ldots, x^{i_{j-1}}, \bm{x}_t).
\end{equation}
We can therefore sample the subsequence one position at a time by conditioning on previously sampled positions and repeatedly applying \emph{marginal inference}.
Specifically, given a set $S \subseteq \{1, \ldots, N\}$ of conditioned positions with values $\bm{x}_S := \{x^j\}_{j \in S}$, we compute the marginal $p(x^i | \bm{x}_S, \bm{x}_t)$ at every unconditioned position $i \notin S$.
Sampling the full sequence then proceeds by initializing $S = \emptyset$ and iteratively (i) performing marginal inference, (ii) selecting a position $i \notin S$, (iii) sampling $x^i \sim p(x^i | \bm{x}_S, \bm{x}_t)$, and (iv) adding $i$ to $S$. The position selection in step~(ii) can follow any logit-based strategy (e.g., top probability-based~\citep{zheng2024topprobabilitystrategy} or entropy-based~\citep{ye2025dream} selection). The number of tokens sampled in a single step can be either fixed or determined by an adaptive sampler such as EB-Sampler~\citep{benhamu2025ebsampler}.
We now describe how to perform marginal inference under each decomposition.

\paragraph{CPD.}
For each position $j$ and rank component $\alpha$, define the \emph{conditioning factor}
\begin{equation}
\bar{p}^{(\alpha, j)}_\theta =
\begin{cases}
    p^{(\alpha, j)}_\theta(x^j | \bm{x}_t) & \text{if } j \in S, \\
    1 & \text{otherwise}.
\end{cases}
\end{equation}
Substituting into the CPD form (\cref{eq:cpd}), for any $i \notin S$,
\begin{equation}
p(x^i | \bm{x}_S, \bm{x}_t) = \frac{p(x^i, \bm{x}_S | \bm{x}_t)}{p(\bm{x}_S | \bm{x}_t)} \propto p(x^i, \bm{x}_S | \bm{x}_t) = \sum_{\alpha=1}^{r} w^{(\alpha)}_\theta(\bm{x}_t)\, p^{(\alpha, i)}_\theta(x^i | \bm{x}_t) \prod_{j=1}^{N} \bar{p}^{(\alpha, j)}_\theta.
\end{equation}
Since $\bar{p}^{(\alpha, i)}_\theta = 1$ for $i \notin S$, the product $\prod_{j=1}^{N} \bar{p}^{(\alpha, j)}_\theta$ does not depend on $i$ and can therefore be precomputed once and reused across all positions, enabling parallel evaluation of $p(x^i, \bm{x}_S | \bm{x}_t)$ for every $i \notin S$. The marginal $p(x^i | \bm{x}_S, \bm{x}_t)$ is then obtained by normalizing $p(x^i, \bm{x}_S | \bm{x}_t)$ over the vocabulary. (Division by $p(\bm{x}_S | \bm{x}_t)$ is also possible but unnecessary and less numerically stable.)

\paragraph{TTD.}
Analogously, for each position $j$, define the \emph{contracted core}
\begin{equation}
\bar{\bm{G}}_j =
\begin{cases}
    \bm{G}_{j,\theta}(x^j | \bm{x}_t) & \text{if } j \in S, \\
    \displaystyle \sum_{x^j = 1}^{V} \bm{G}_{j,\theta}(x^j | \bm{x}_t) & \text{otherwise}.
\end{cases}
\end{equation}
When $j \notin S$, $\bar{\bm{G}}_j$ does not depend on the chain-rule conditioning $\bm{x}_S$ and can be cached across queries. For any $i \notin S$,
\begin{equation}
p(x^i, \bm{x}_S | \bm{x}_t) = \tfrac{1}{r}\, \bm{1}_r^{\top}\, \bar{\bm{G}}_1 \cdots \bar{\bm{G}}_{i-1}\, \bm{G}_{i,\theta}(x^i | \bm{x}_t)\, \bar{\bm{G}}_{i+1} \cdots \bar{\bm{G}}_N\, \bm{1}_r.
\end{equation}
The left-cumulative products $\bar{\bm{G}}_1 \cdots \bar{\bm{G}}_{i-1}$ and right-cumulative products $\bar{\bm{G}}_{i+1} \cdots \bar{\bm{G}}_N$ for all $i$ can be obtained via parallel prefix sums. Normalizing $p(x^i, \bm{x}_S | \bm{x}_t)$ over the vocabulary then yields $p(x^i | \bm{x}_S, \bm{x}_t)$, again avoiding the explicit computation of $p(\bm{x}_S | \bm{x}_t)$.

\subsubsection{Special Case: Predetermined Sampling Positions}
\label{sec:sampling_special_case}

While the method above samples one position at a time, in many settings---random ordering, fixed orderings such as left-to-right --- the positions to be sampled in the current step are determined before any computation, allowing them to be sampled \emph{jointly}. Let $T \subseteq \{1, \ldots, N\}$ denote the masked positions to be sampled jointly in the same step, with values $\bm{x}_T := \{x^j\}_{j \in T}$. For CPD:
\begin{equation}
p(\bm{x}_T | \bm{x}_t) = \sum_{\alpha=1}^{r} w^{(\alpha)}_\theta(\bm{x}_t) \prod_{j \in T} p^{(\alpha, j)}_\theta(x^j | \bm{x}_t)
\end{equation}
Sampling therefore requires evaluating $p^{(\alpha, j)}_\theta(x^j | \bm{x}_t)$ only at positions $j \in T$.

With TTD, positions $j \notin T$ enter $p(\bm{x}_T | \bm{x}_t)$ only through the contracted cores $\bar{\bm{G}}_j$. By absorbing each $\bar{\bm{G}}_j$ into a neighboring core, the remaining cores form a TT representation over only the $|T|$ positions in $T$. On this we can run a highly optimized version of the general TTD method, by caching cumulative products and computing $p(x^i | \bm{x}_S, \bm{x}_t)$ for only the next immediate value of $i$.

The reduction above assumes access to $\bar{\bm{G}}_j$ for each $j \notin T$. Computing it from the definition by outputting $\bm{G}_{j,\theta}(x^j | \bm{x}_t)$ for all $V$ values of $x^j$ and summing is computationally wasteful. We sidestep this overhead by introducing an additional head $\bar{\bm{G}}_{j,\theta}(\bm{x}_t)$ that directly produces the contracted core, trained to be consistent with $\bm{G}_{j,\theta}(x^j | \bm{x}_t)$:
\begin{equation}
    \mathcal{L}_{\text{C}} = \frac{1}{N} \sum_{j=1}^{N} \Bigl\lVert \bar{\bm{G}}_{j,\theta}(\bm{x}_t) - \text{stopgrad}(\sum_{x^j} \bm{G}_{j,\theta}(x^j | \bm{x}_t)) \Bigr\rVert_F^2
    \label{eqn:l_c}
\end{equation}
then optimize
\begin{equation}
    \mathcal{L} = \mathcal{L}_{\text{NLL}} + \mathcal{L}_{\text{C}}.
\end{equation}

\subsection{Extension to Latent-Variable Approaches}
\label{sec:vadd-extension}

Our framework is flexible: beyond a base MDM, it can also be combined with latent-variable approaches. We illustrate this with Variational Autoencoding Discrete Diffusion (VADD)~\citep{xie2026vadd}, a few-step MDM framework based on a latent variable model. Integrating our explicit modeling of $p_{\bm\theta}(\bm{x} | \bm{x}_t)$ into VADD yields the training loss
\begin{equation}
\widehat{\mathcal{L}}_{\lambda}(\bm{x}_0;\bm{\theta},\bm{\phi}) = \int_{0}^{1} \mathbb{E}_{q(\bm{x}_t|\bm{x}_0)}\mathbb{E}_{r_{\bm{\phi}}(\bm{z}|\bm{x}_0, \bm{x}_t)}\frac{-\alpha_t'}{1-\alpha_t} \left[\log p_{\bm{\theta}}(\bm{x}_0|\bm{x}_t,\bm{z})-\lambda\log\left(\frac{r_{\bm{\phi}}(\bm{z}|\bm{x}_0, \bm{x}_t)}{p(\bm{z})}\right)\right] \mathrm{d} t,
\end{equation}
with the sampling procedure
\[
\bm{z} \sim p(\bm{z}), \qquad \bm{x}_{t_{i-1}} \sim p_{\bm\theta}(\bm{x}_{t_{i-1}}|\bm{x}_{t_i},\bm{z}),
\]
where our tensor decomposition parameterizes both $p_{\bm{\theta}}(\bm{x}_0|\bm{x}_t,\bm{z})$ and $p_{\bm{\theta}}(\bm{x}_{t_{i-1}}|\bm{x}_{t_i},\bm{z})$. Results for this combination are reported in \Cref{sec:results_text_generation}.

\subsection{Fine-Tuning Pretrained MDMs and Implementation Details}

Our tensor-decomposition framework can be integrated into pretrained MDMs via fine-tuning. We describe the necessary architectural modifications and other implementation details below.

\begin{figure}
  \centering
  \includegraphics[width=1\linewidth]{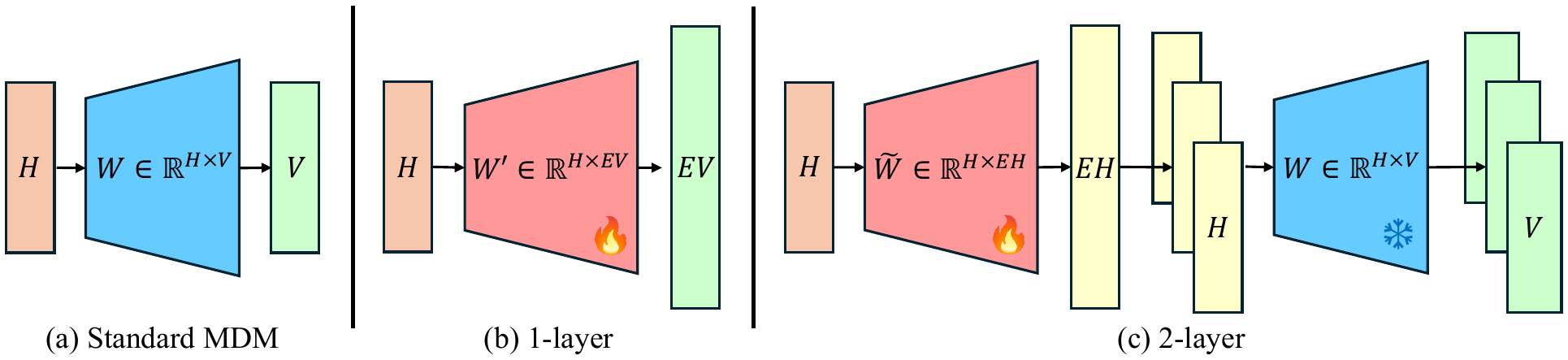}
  \caption{Architectural modifications for fine-tuning MDMs.}
  \label{fig:architectures}
\end{figure}

\paragraph{Architecture.}
A standard MDM produces $NV$ logits, but tensor-decomposed forms require $EVN$, where $E = r$ (CPD; plus $r$ rank weights) or $E = r^2$ (TTD). Our \emph{1-layer} method replaces the final head $\bm{W} \in \mathbb{R}^{H \times V}$ with $\bm{W}' \in \mathbb{R}^{H \times EV}$, which is costly for large $V$. Following \citet{basharin2025fasterlanguagemodelsbetter}, the \emph{2-layer} method instead inserts $\widetilde{\bm{W}} \in \mathbb{R}^{H \times EH}$ before the final head, then applies $\bm{W}$ to each of the $E$ outputs, which significantly reduces the parameter count since $H \ll V$ in language models. Schematics of different architectures are shown in~\cref{fig:architectures}.

\paragraph{Initialization.}
We initialize the added weights to preserve pretrained marginal predictions: $\bm{W}'$ as $E$ copies of $\bm{W}$ (1-layer), or $\widetilde{\bm{W}}$ as $E$ stacked identity matrices with $\bm{W}$ frozen (2-layer). Small Gaussian noise is added to break rank-symmetry.

\paragraph{Masked-token-only modeling.} 
Let $\mathcal{M} := \{i : x_t^i = m\}$ denote the set of masked token indices. During the forward process, unmasked tokens remain unchanged (\ie $x_t^i = x^i$ for $i \notin \mathcal{M}$), so the likelihood factorizes as $p_\theta(\bm{x} | \bm{x}_t) = p_\theta(\bm{x}_\mathcal{M} | \bm{x}_t) \prod_{i \notin \mathcal{M}} \delta_{x_t^i}(x^i)$. We therefore restrict network forward passes to predicting $p_\theta(\bm{x}_\mathcal{M} | \bm{x}_t)$ over masked tokens only for efficiency. The NLL loss is normalized by $|\mathcal{M}|$ accordingly.

\section{Experimental Results}
\label{sec:results}

\subsection{Text Generation}
\label{sec:results_text_generation}

We demonstrate the application of our methods on both MDLM~\citep{sahoo2024mdlm} and VADD~\citep{xie2026vadd}. For MDLM~\citep{sahoo2024mdlm}, we use the official checkpoint trained on OpenWebText (OWT) \citep{gokaslan2019owt} as the base model, and for VADD \citep{xie2026vadd} we train a base model from scratch for both OWT and LM1B~\citep{chelba2014lm1b}. These models are augmented with the 2-layer architecture~\citep{basharin2025fasterlanguagemodelsbetter} before fine-tuning, with $r_{\text{CPD}}=16, r_{\text{TTD}}=4$. For training and sampling with VADD and other baselines~\citep{xu2025edlm,hayakawa2025di4c}, we use their official implementation, while fixing the numerical issue in categorical sampling reported by~\citet{zheng2025maskeddiffusionmodelssecretly}, using 64-bit precision for logits. Additional details on training and sampling can be found in~\app~\ref{app:experiment_details_unconditional_text_generation}.

\begin{table}[!t]
  \caption{Generative perplexity (GPT-2) on OpenWebText~\citep{gokaslan2019owt}.}
  \label{tab:owt-genppl}
  \begin{threeparttable}
  \resizebox{\columnwidth}{!}{%
  \begin{tabular}{l|c|c|c}
  \toprule
    \text{Method} & \text{Timesteps} & \text{Generative Perplexity} ($\downarrow$) & \text{Entropy} \\
    \toprule
    \multicolumn{4}{c}{\textit{MDLM-based}} \\
    \midrule
    MDLM & 8 / 16 / 32 / 64 / 128 & 840.57 / 347.10 / 195.09 / 146.09 / 122.08 & 7.77 / 7.67 / 7.60 / 7.58 / 7.56 \\
    EDLM & 8 / 16 / 32 / 64 / 128 & 891.18 / 367.48 / 212.02 / 153.79 / 129.38 & 7.83 / 7.73 / 7.66 / 7.62 / 7.60 \\
    Di4C & 8 / 16 / 32 / 64 / 128 & 832.52 / 360.89 / 207.50 / 152.63 / 127.31 & 7.79 / 7.70 / 7.63 / 7.60 / 7.57 \\
    MDLM+CPD (Ours) & 8 / 16 / 32 / 64 / 128 & 848.87 / 358.25 / 205.46 / 148.10 / 126.85 & 7.81 / 7.71 / 7.65 / 7.60 / 7.60 \\
    MDLM+TTD (Ours) & 8 / 16 / 32 / 64 / 128 & \textbf{636.98} / \textbf{295.43} / \textbf{183.43} / \textbf{140.97} / \textbf{119.20} & 7.75 / 7.66 / 7.60 / 7.57 / 7.54 \\
    \midrule
    \multicolumn{4}{c}{\textit{SEDD-based}} \\
    \midrule
    DCD & 8 / 16 / 32 / 64 / 128 & 252.11 / 240.75 / 201.75 / 189.09 / 176.74 & 8.65 / 8.76 / 8.83 / 8.89 / 8.89 \\
    \midrule
    \multicolumn{4}{c}{\textit{VADD-based}} \\
    \midrule
    VADD & 8 / 16 / 32 / 64 / 128 & 331.69 / 204.31 / 151.84 / 132.09 / 123.09 & 7.05 / 7.02 / 6.98 / 6.98 / 6.97 \\
    VADD+CPD (Ours) & 8 / 16 / 32 / 64 / 128 & 299.17 / 180.72 / 135.97 / 118.11 / 108.57 & 6.93 / 6.90 / 6.86 / 6.86 / 6.84 \\
    VADD+TTD (Ours) & 8 / 16 / 32 / 64 / 128 & \textbf{223.19} / \textbf{153.85} / \textbf{122.34} / \textbf{109.80} / \textbf{101.70} & 6.82 / 6.80 / 6.78 / 6.78 / 6.76 \\
    \bottomrule
  \end{tabular}}
  \end{threeparttable}
\end{table}

\begin{table}[!t]
  \caption{Generative perplexity (GPT-2) on LM1B~\citep{chelba2014lm1b}.}
  \label{tab:lm1b-genppl}
  \begin{threeparttable}
  \resizebox{\columnwidth}{!}{%
  \begin{tabular}{l|c|c|c}
  \toprule
    \text{Method} & \text{Timesteps} & \text{Generative Perplexity} ($\downarrow$) & \text{Entropy} \\
    \toprule
    VADD & 1 / 2 / 4 / 8 / 16 & 1096.39 / 664.16 / 414.02 / 299.39 / 251.03 & 6.94 / 6.90 / 6.90 / 6.93 / 6.93 \\
    VADD+CPD (Ours) & 1 / 2 / 4 / 8 / 16 & 1099.93 / 659.87 / 419.69 / 298.66 / 249.77 & 6.95 / 6.90 / 6.90 / 6.93 / 6.93 \\
    VADD+TTD (Ours) & 1 / 2 / 4 / 8 / 16 & \textbf{850.64} / \textbf{532.43} / \textbf{369.56} / \textbf{284.86} / \textbf{247.40} & 6.99 / 6.93 / 6.94 / 6.95 / 6.96 \\
    \bottomrule
  \end{tabular}}
  \end{threeparttable}
\end{table}

\begin{table}[!t]
  \caption{Rank size ablation on OpenWebText~\citep{gokaslan2019owt}, measured by generative perplexity (GPT-2).}
  \label{tab:owt-genppl-rank}
  \begin{threeparttable}
  \resizebox{\columnwidth}{!}{%
  \begin{tabular}{l|c|c|c}
  \toprule
    \text{Method} & \text{Timesteps} & \text{Generative Perplexity} ($\downarrow$) & \text{Entropy} \\
    \toprule
    MDLM & 8 / 16 / 32 / 64 / 128 & 840.57 / 347.10 / 195.09 / 146.09 / 122.08 & 7.77 / 7.67 / 7.60 / 7.58 / 7.56 \\
    MDLM+TTD (r=2) & 8 / 16 / 32 / 64 / 128 & 703.02 / 304.57 / 185.11 / \textbf{140.47} / 120.74 & 7.75 / 7.63 / 7.59 / 7.55 / 7.55 \\
    MDLM+TTD (r=3) & 8 / 16 / 32 / 64 / 128 & 664.63 / 299.72 / 187.73 / 140.82 / 123.08 & 7.78 / 7.66 / 7.61 / 7.58 / 7.58 \\
    MDLM+TTD (r=4) & 8 / 16 / 32 / 64 / 128 & \textbf{636.98} / \textbf{295.43} / \textbf{183.43} / 140.97 / \textbf{119.20} & 7.75 / 7.66 / 7.60 / 7.57 / 7.54 \\
    \bottomrule
  \end{tabular}}
  \end{threeparttable}
\end{table}

We sample 1024 samples from each using random ordering, then measure the generative perplexity with gpt2-large \citep{radford2019gpt2}, and entropy of the generated sequences as a measure of diversity. The sample length is 1024 for OWT~\citep{gokaslan2019owt} and 128 for LM1B \citep{chelba2014lm1b}.

Results on OWT~\citep{gokaslan2019owt} show that TTD significantly improves over base MDLM~\citep{sahoo2024mdlm}, reducing generative perplexity by 24.22\% and 14.88\% in the challenging 8- and 16-step regimes, while CPD yields modest to no improvement (\cref{tab:owt-genppl}, Top). A similar trend holds for VADD+TTD on the same dataset (\cref{tab:owt-genppl}, Bottom) and on LM1B~\citep{chelba2014lm1b}, where large perplexity gaps persist across all sampling steps (\cref{tab:lm1b-genppl}). We conjecture that the gain of TTD on VADD stems from two directions of joint probability modeling: VADD captures high-level semantics via a latent variable, while TTD models low-level dependencies between neighboring tokens through lightweight fine-tuning with the objective in~\cref{sec:sampling_special_case}.

We further analyze the proposed method along two axes: rank size impact and runtime overhead. We begin by ablating the rank size using MDLM~\citep{sahoo2024mdlm} trained on OWT~\citep{gokaslan2019owt}, by varying the rank size from $r=2$ to $r=4$. The results are summarized in~\cref{tab:owt-genppl-rank}. As shown, generative perplexity largely improves as rank increases, validating that greater expressibility of the decomposition helps in capturing the joint distribution.

\begin{wraptable}{r}{0.55\columnwidth}
  \caption{Sampling time on OpenWebText~\citep{gokaslan2019owt}, measured with 128 timesteps.}
  \label{tab:owt-time}
  \centering
  {\scriptsize
  \begin{tabular*}{0.55\columnwidth}{@{\extracolsep{\fill}}l|c|l|c}
  \toprule
    \text{Method} & \text{Sec/Seq} ($\downarrow$) & \text{Method} & \text{Sec/Seq} ($\downarrow$) \\
    \midrule
    MDLM & 2.39 & DCD & 550.62 \\
    EDLM & 6.71 & VADD & 3.03 \\
    Di4C & 3.63 & VADD+CPD (Ours) & 2.48 \\
    MDLM+CPD (Ours) & 2.37 & VADD+TTD (Ours) & 3.08 \\
    MDLM+TTD (Ours) & 2.96 &  &  \\
    \bottomrule
  \end{tabular*}}
\end{wraptable}

Due to the large $V$ in language models, na\"{i}vely applying our methods can impose a significant overhead. However, the techniques discussed in~\cref{sec:sampling_special_case} substantially reduce this cost. 
To demonstrate this, we measure sampling time across methods at 128 timesteps, and report the results in~\cref{tab:owt-time}. Notably, on VADD+TTD, the best-performing approach on OWT~\citep{gokaslan2019owt} by generative perplexity, our method is only 1.7\% slower compared to the original VADD implementation, demonstrating practical efficiency. Moreover, we stress that the overhead stems from the size of output tensors rather than model size, and thus does not scale with model depth; we expect it to be negligible for larger diffusion language models.

\subsection{Molecule Generation}
\label{sec:results_molecule_generation}

We evaluate on the molecule generation task using QM9~\citep{blum2009qm9_1,rupp2012qm9_2}, with MDLM~\citep{sahoo2024mdlm}, parameterized using DiT~\citep{peebles2023dit}, as the base model. Given the smaller vocabulary size ($V=40$) relative to text generation, a 1-layer architecture suffices, with $r_{\text{CPD}}=64, r_{\text{TTD}}=8$. Training and sampling details are described in~\app~\ref{app:experiment_details_molecule_generation}.

For the DiT architecture, we modify the rotary positional encoding to apply the rotary matrices to values as well. We stress that this is crucial for any MDM-like sampling methods. Without such a modification, the first sampling step collapses entirely for both the baseline MDLM and our methods due to the uniformity of the inputs in every position. The necessity of such a modification is proved in~\app~\ref{app:pitfall_of_rope_in_masked_diffusion}.

\begin{figure}
  \centering
  \includegraphics[width=1\linewidth]{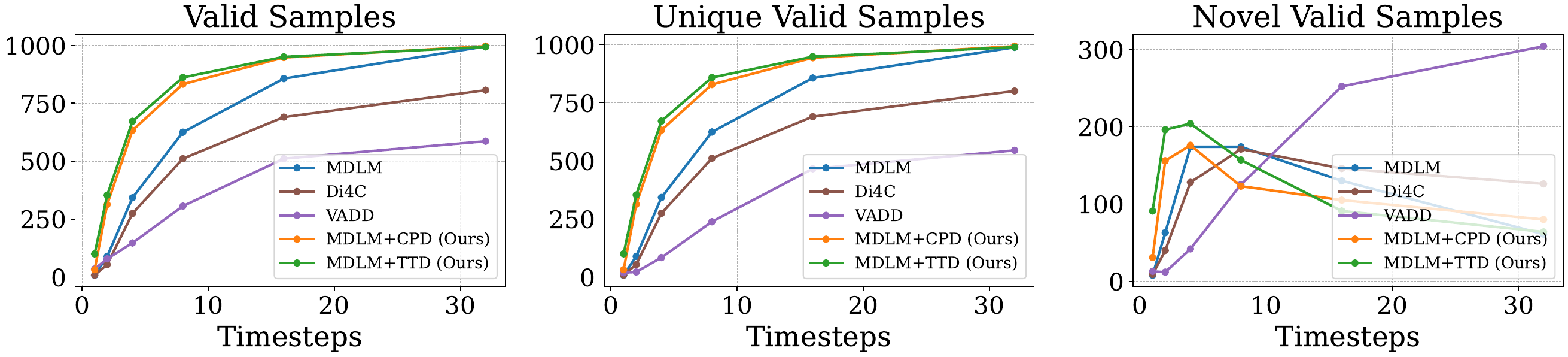}
  \caption{Molecule generation on QM9~\citep{blum2009qm9_1,rupp2012qm9_2}. The proposed method consistently achieves higher validity than baselines, including few-step regimes.}
  \label{fig:qm9_comparison}
\end{figure}

We evaluate each method by sampling 1024 molecules with random ordering across varying numbers of sampling steps, then compute validity, and uniqueness and novelty among valid samples. These metrics are summarized in~\cref{fig:qm9_comparison}. We find that our methods bring significant improvements despite small increases in model size. Additional experiments, including sampling strategy comparisons and rank size ablations, are provided in~\app~\ref{app:additional_experiments_molecule_generation}. We observe that, as predicted from our discussion in~\cref{sec:method}, TTD outperforms CPD of the same output size, and that this gap is the largest in left-to-right generation, where the structural bias of TTD can be exploited to the fullest extent.
\section{Conclusion}
We presented the first framework for explicit joint distribution modeling in discrete diffusion, parameterizing the conditional clean distribution $p_{\bm\theta}(\bm{x}|\bm{x}_t)$ as a low-rank tensor via CPD or TTD and thus generalizing standard MDMs. Our analysis identifies a structural bias of TTD toward dependencies between nearby tokens, making it particularly well-suited to sequential data. Through an efficient sampling procedure and a lightweight fine-tuning recipe, the framework integrates with any base MDM or latent-variable method and substantially improves few-step generation on both text and molecule benchmarks. 
\section*{Societal Impacts}
This work targets efficiency improvements for discrete diffusion generation. On the positive side, faster sampling can reduce the computational cost and energy use of sequential generation, and may accelerate beneficial downstream applications such as molecular design and biological sequence modeling. As with any advance in generative model efficiency, there is also potential for misuse (e.g., faster generation of misleading textual content), though such risks are inherent to the broader generative modeling paradigm rather than introduced by our framework.
\section*{Limitations and Future Work}
A fundamental challenge in discrete diffusion modeling is that the joint distribution $p_{\bm\theta}(\bm{x}|\bm{x}_t)$ over $N$ tokens corresponds to an $N$-dimensional tensor with $V^N$ entries. For arbitrary target distributions this tensor can be of prohibitively high rank, so any tractable parameterization, including ours, must rely on a low-rank approximation, and arbitrarily increasing the rank in CPD or TTD is infeasible in practice. Nonetheless, our experiments show that even low ranks yield substantial few-step improvements across both text and molecule generation, indicating that low-rank structure suffices in the regime where parallelization bias is most severe. As future work, we plan to explore richer tensor formats, such as Tucker decomposition, which may capture dependency structures beyond the locality bias of TTD and offer alternative trade-offs between rank and expressivity.

\bibliographystyle{plainnat}
\bibliography{refs.bib}

\newpage

\appendix
\crefname{appendix}{Appendix}{Appendices}
\Crefname{appendix}{Appendix}{Appendices}
\section*{Appendix}
\section{Normalization Constraints}
\label{app:normalization_constraints}

In \Cref{sec:method}, we introduced normalization constraints for both decompositions. Here, we prove that (i) these constraints ensure $p_\theta(\bm{x} \vert \bm{x}_t)$ is a valid probability mass function (\ie sums to 1 over $V^N$), and (ii) they do not reduce expressibility: any normalized tensor representable by the given decomposition is also representable under the constraints. We first prove the claim for Canonical Polyadic Decomposition (CPD), then for Tensor Train Decomposition (TTD).

\subsection{Canonical Polyadic Decomposition (CPD)}

For CPD, recall that the constraints are given as:
\begin{equation}
    \sum_{\alpha=1}^{r} w^{(\alpha)}_\theta(\bm{x}_t) = 1, \qquad \sum_{x^i=1}^{V} p^{(\alpha, i)}_\theta(x^i | \bm{x}_t) = 1.
\end{equation}
We check (i) by:

\begin{multline}
\sum \limits_{\bm x} p_\theta(\bm{x} | \bm{x}_t)= \sum \limits_{\bm x} \sum_{\alpha=1}^{r} w^{(\alpha)}_\theta(\bm{x}_t) \prod_{i=1}^{N} p^{(\alpha, i)}_\theta(x^i | \bm{x}_t)=  \\ \sum \limits_{\alpha=1}^r w^{(\alpha)}_\theta(\bm{x}_t) \prod \limits_{i=1}^N \sum \limits_{x^i=1}^V p^{(\alpha, i)}_\theta(x^i | \bm{x}_t)=\sum \limits_{\alpha=1}^r w^{(\alpha)}_\theta(\bm{x}_t) \prod \limits_{i=1}^N 1 = 1.
\end{multline}
We prove (ii) by starting with a representation of a normalized tensor, then converting it to a constraint-satisfying representation of the same tensor. Given $w^{(\alpha)}_\theta(\bm{x}_t)$ and $p^{(\alpha, i)}_\theta(x^i | \bm{x}_t)$, for every $\alpha$, we divide each $p^{(\alpha, i)}_\theta(x^i | \bm{x}_t)$ by $\sum \limits_{x^i=1}^V p^{(\alpha, i)}_\theta(x^i | \bm{x}_t)$, and multiply $w^{(\alpha)}_\theta(\bm{x}_t)$ by the same amount, keeping the tensor unchanged. Then we have $\sum \limits_{x^i=1}^V p^{(\alpha, i)}_\theta(x^i | \bm{x}_t)=1$. And by
\[1 = \sum \limits_{\bm x} p(\bm x)= \sum \limits_{\alpha=1}^r w^{(\alpha)}_\theta(\bm{x}_t) \prod \limits_{i=1}^N \sum \limits_{x^i=1}^V p^{(\alpha, i)}_\theta(x^i | \bm{x}_t)=\sum \limits_{\alpha=1}^r w^{(\alpha)}_\theta(\bm{x}_t)\]
we get $\sum \limits_{\alpha=1}^r w^{(\alpha)}_\theta(\bm{x}_t)=1$. It represents the same tensor, and satisfies the constraints.

\subsection{Tensor-Train Decomposition (TTD)}

For TTD, recall that the constraint is given as:
\begin{equation}
\sum_{x^i, k} \bm{G}_{i,\theta}(x^i, j, k | \bm{x}_t) = 1, \quad \forall i,j.
\end{equation}
Similar to our proof for CPD, we begin by verifying (i):
\begin{multline}
\sum \limits_{\bm x} p_\theta(\bm{x} | \bm{x}_t)=\sum \limits_{\bm x} \tfrac{1}{r} \bm{1}_r^T \bm{G}_{1,\theta}(x^1 | \bm{x}_t) \bm{G}_{2,\theta}(x^2 | \bm{x}_t) ... \bm{G}_{N,\theta}(x^N | \bm{x}_t) \bm{1}_r \\ =\tfrac{1}{r} \bm{1}_r^T \sum \limits_{x^1=1}^V \bm{G}_{1,\theta}(x^1 | \bm{x}_t) ... \sum \limits_{x^N=1}^V \bm{G}_{N,\theta}(x^N | \bm{x}_t) \bm{1}_r.
\end{multline}
Since by constraint $\sum \limits_{x^i=1}^V \bm{G}_{i,\theta}(x^i | \bm{x}_t) \bm{1}_r = \bm{1}_r$, the summations collapse:
\begin{equation}
\sum \limits_{\bm x} p_\theta(\bm{x} | \bm{x}_t)=\tfrac{1}{r} \bm{1}_r^T \sum \limits_{x^1=1}^V \bm{G}_{1,\theta}(x^1| \bm{x}_t) ... \sum \limits_{x^{N-1}=1}^V \bm{G}_{N-1,\theta}(x^{N-1} | \bm{x}_t) \bm{1}_r = ... = \tfrac{1}{r} \bm{1}_r^T \bm{1}_r = 1
\end{equation}
We prove (ii) in the same way as CPD. Given the cores for a normalized tensor, for position $i$, for every $j$, we divide each $\bm{G}_{i,\theta}(x^i,j,k | \bm{x}_t)$ by $\sum \limits_{x^i, k} \bm{G}_{i,\theta}(x^i,j,k | \bm{x}_t)$, and multiply the same term to each $\bm{G}_{i-1,\theta}(x^{i-1},l,j | \bm{x}_t)$, to keep the tensor unchanged. After this, each $\bm{G}_{i,\theta}(x^i | \bm{x}_t)$ satisfies the constraints at the cost of disrupting $\bm{G}_{i-1,\theta}(x^{i-1} | \bm{x}_t)$. This is applied for $i=N,\dots,2$ in decreasing order, leaving all but $\bm{G}_{1,\theta}(x^1 | \bm{x}_t)$ well-constrained.
For $\bm{G}_{1,\theta}(x^1 | \bm{x}_t)$, we set $\bm{G}_{1,\theta}(x^1, l, j | \bm{x}_t)=\tfrac{1}{r} \sum \limits_{l} \bm{G}_{1,\theta}(x^1, l, j | \bm{x}_t)$, i.e. we run in-place column-wise mean for each matrix. This preserves the tensor, since $\bm{1}_r^T \bm{G}_{1,\theta}(x^1 | \bm{x}_t)$ remains the same. Also it gives the final constraint, because
\[1=\sum \limits_{\bm x} p_\theta(\bm{x} | \bm{x}_t)=\tfrac{1}{r} \bm{1}_r^T \sum \limits_{x^1=1}^V \bm{G}_{1,\theta}(x^1| \bm{x}_t) ... \sum \limits_{x^N=1}^V \bm{G}_{N,\theta}(x^N| \bm{x}_t) \bm{1}_r=\tfrac{1}{r} \bm{1}_r^T \sum \limits_{x^1=1}^V \bm{G}_{1,\theta}(x^1| \bm{x}_t) \bm{1}_r\]
is equivalent to $\sum \limits_{x^1, j, k} \bm{G}_{1,\theta}(x^1,j,k | \bm{x}_t)=r$, thus after running the column-wise mean $\sum \limits_{x^1, k} \bm{G}_{1,\theta}(x^1, j, k | \bm{x}_t)=1$ for every $j$. It satisfies all constraints.

\section{An Example Demonstrating TTD’s Advantage}
\label{app:example_of_ttd_advantage}

As claimed in \Cref{sec:ttd}, we provide a simple distribution with neighbor-only dependence that can be modeled with a low TT-rank but not with a low CP-rank. Consider a tensor
\[\bm P=\underbrace{I_2 \otimes I_2 \otimes ... \otimes I_2}_{N-\text{times}} / 2^N,\]
representing a PMF defined over $2N$-sequences consisting of binary tokens where the paired neighbors share identical values (\eg $1100111100$).
We show that $\bm P$ has $r_{\text{CPD}}=2^N$, but $r_{\text{TTD}}=2$.

We first show that $r_{\text{CPD}} \leq 2^N$. Note that $\bm P$ can be written as a sum of scaled Dirac deltas, with each indicating a possible sequence, while being rank-1 tensors. There are $2^N$ of these, so the rank is at most $2^N$.

Next, we show that $r_{\text{CPD}} \geq 2^N$ by using the technique of matrix flattening. Sending the odd indices to the row, and even indices to the column, and we obtain a matrix of size $2^N \times 2^N$. From the definition of $\bm P$, this becomes the scaled identity matrix.
This matrix has rank $2^N$, and since the tensor rank is always greater than or equal to the matrix rank of any of its flattenings, we get $r_{\text{CPD}} \geq 2^N$.

On the other hand $r_{\text{TTD}} \leq 2$: We construct an explicit representation with $r_{\text{TTD}}=2$ :
\[\bm G_{2k-1}(0)=\begin{bmatrix}1/2 & 0 \\ 1/2 & 0\end{bmatrix},
\bm G_{2k-1}(1)=\begin{bmatrix}0 & 1/2 \\ 0 & 1/2\end{bmatrix}\]
\[\bm G_{2k}(0)=\begin{bmatrix}1/2 & 1/2 \\ 0 & 0\end{bmatrix},
\bm G_{2k}(1)=\begin{bmatrix}0 & 0 \\ 1/2 & 1/2\end{bmatrix}\]
then
\[\bm P(\nu_1, \ldots, \nu_{2N})=\tfrac{1}{r} \bm{1}_r^T \bm G_1(\nu_1) \bm G_2(\nu_2) ... \bm G_{2N-1}(\nu_{2N-1}) \bm G_{2N}(\nu_{2N}) \bm{1}_r\]

$r_{\text{TTD}} \geq 2$: $r_{\text{TTD}}=1$ is equivalent to being a rank-1 tensor, which is not the case since $r_{\text{CPD}} \geq 2^N$.

Thus $r_{\text{CPD}}=2^N, r_{\text{TTD}}=2$.  $ \square$

\section{A Pitfall of RoPE in Masked Diffusion}
\label{app:pitfall_of_rope_in_masked_diffusion}

Rotary Position Embedding (RoPE) \citep{su2024roformer} incorporates positional information into the attention layer by "rotating" queries and keys by
\[q_m=R^d_{\Theta, m} W_q x_m, k_n=R^d_{\Theta, n} W_k x_n\]
where $R^d_{\Theta, m}$ and $R^d_{\Theta, n}$ are appropriately defined rotary matrices. (Here we write position indices in subscripts to match \citet{su2024roformer}.)

This method, however, should be used with care on MDMs. In the very first step of MDM generation, the input is a fully-masked sequence. Since default RoPE neither changes the token embeddings nor the values in attention layers, it loses its ability to distinguish between positions.

If the initial $x_m=x$ is uniform across position, $v_m=W_v x_m=W_v x=v$ is also uniform, and thus the output of attention layer 
\begin{align}
o_m=\sum \limits_{n=1}^N a_{m,n}v_n=v\sum \limits_{n=1}^N a_{m,n}=v    
\end{align}
also becomes uniform regardless of the attention weights $a$. All the other components, such as normalizing layers or dense layers, are applied independently to each position, maintaining this unintended uniformity. Indeed, we confirm that under this formulation, the outputted logits of the model are identical for every position in the first step.

Although this issue is less pronounced in settings such as conditional generation (where the first step is not fully masked), long sequence generation (where the impact of an incorrectly chosen first token is negligible), or datasets where true marginal probabilities are similar across positions, we observed noticeable performance degradation in our QM9~\citep{blum2009qm9_1,rupp2012qm9_2} experiments, where the model performs unconditional generation of short sequences. This motivated a fix: applying RoPE rotary matrices to values in addition to queries and keys in attention. Note that we retain the previous behavior in~\Cref{sec:results_text_generation} for compatibility with the pretrained model.

\section{Experiment Details}

\subsection{Text Generation}
\label{app:experiment_details_unconditional_text_generation}

For OWT~\citep{gokaslan2019owt} and LM1B~\citep{chelba2014lm1b}, we matched the training time for each method. Each resulted in a total training time of approximately 10 hours on four NVIDIA A6000 GPUs.

For MDLM~\citep{sahoo2024mdlm} we fine-tuned the base model with global batch size 64, constant learning rate $1 \times 10^{-5}$ with Adam for 100k steps, and for CPD and TTD we used global batch size 4 for 150k steps.

Di4C~\citep{hayakawa2025di4c} model was trained for 13k steps with the base model as the teacher model, with other hyperparameters set to default values in the released code of global batch size 4 and constant learning rate $3 \times 10^{-5}$ with warmup for 2500 steps with AdamW~\citep{loshchilov2019adamw} optimizer, alongside exponential moving average (EMA) with decay rate of 0.9999.

For EDLM~\citep{xu2025edlm}, we used the checkpoints from MDLM with the AR-based method, and used the default sampling parameter of sampling size $k=2$ but widened the sampling window to $w=1$ to follow the setting outlined in their paper.

For DCD~\citep{liu2025dcd}, we followed the released code and used sedd-medium \citep{lou2024sedd} as the diffusion model and gpt2 \citep{radford2019gpt2} as the AR model with a chunk size of 1 to simulate fully random generation.

For VADD~\citep{xie2026vadd}, we trained the base model with the default values in the released code of global batch size 512 and constant learning rate $3 \times 10^{-4}$ with warmup for 2500 steps with AdamW and EMA with decay rate 0.9999 for 5k steps on latent dimension of 512. VADD and our methods were fine-tuned from the base model in the same setting, with VADD for 1k steps and our methods for 400 steps. For sampling we followed their default setting using fixed latent.

For measuring the sampling times, we generated 64 sequences with 128 timesteps on batch size of 1 for every method and computed the sampling time of a single sequence. Sampling time measurements were conducted on a single NVIDIA PRO6000 GPU.

\subsection{Molecule Generation}
\label{app:experiment_details_molecule_generation}

The base MDLM~\citep{sahoo2024mdlm} was trained with global batch size 8192, constant learning rate $1 \times 10^{-4}$ with Adam~\citep{kingma2015adam} for 50k steps. For MDLM and our methods, we fine-tuned the base model under the same settings for 10k steps.

Di4C~\citep{hayakawa2025di4c} model was trained for 10k steps with our base model as the teacher model, with global batch size 512 and other hyperparameters set to default values in the released code of constant learning rate $3 \times 10^{-5}$ with warmup for 2500 steps with AdamW~\citep{loshchilov2019adamw} and exponential moving average (EMA) with decay rate 0.9999.

VADD~\citep{xie2026vadd} model was trained for 60k steps from scratch with global batch size 2048 and other hyperparameters set to default values in the released code of constant learning rate $3 \times 10^{-4}$ with warmup for 2500 steps with AdamW and EMA with decay rate 0.9999 on latent dimension of 512. For sampling of VADD we also followed their default setting using fixed latent.

Experiments were conducted using four GPUs of either NVIDIA A6000 or PRO6000.

\section{Additional Results}


\subsection{Molecule Generation}
\label{app:additional_experiments_molecule_generation}

\begin{figure}
  \centering
  \includegraphics[width=1\linewidth]{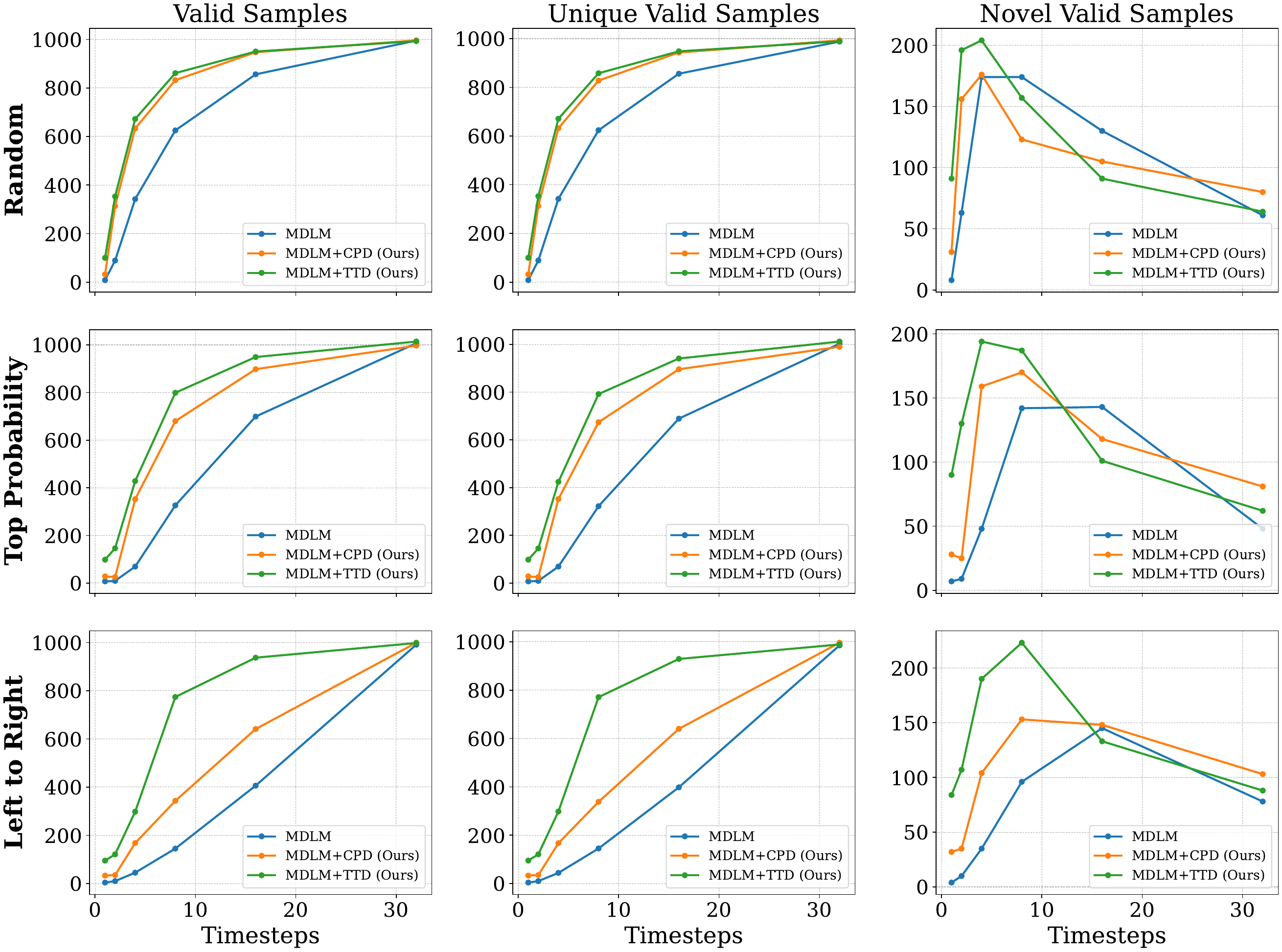}
  \caption{Molecule generation on QM9 \citep{blum2009qm9_1, rupp2012qm9_2} for different sampling strategies. Each row corresponds to random order, top-probability-based order and left-to-right order.}
  \label{fig:QM9-sampling-strategies}
\end{figure}

\begin{figure}
  \centering
  \includegraphics[width=1\linewidth]{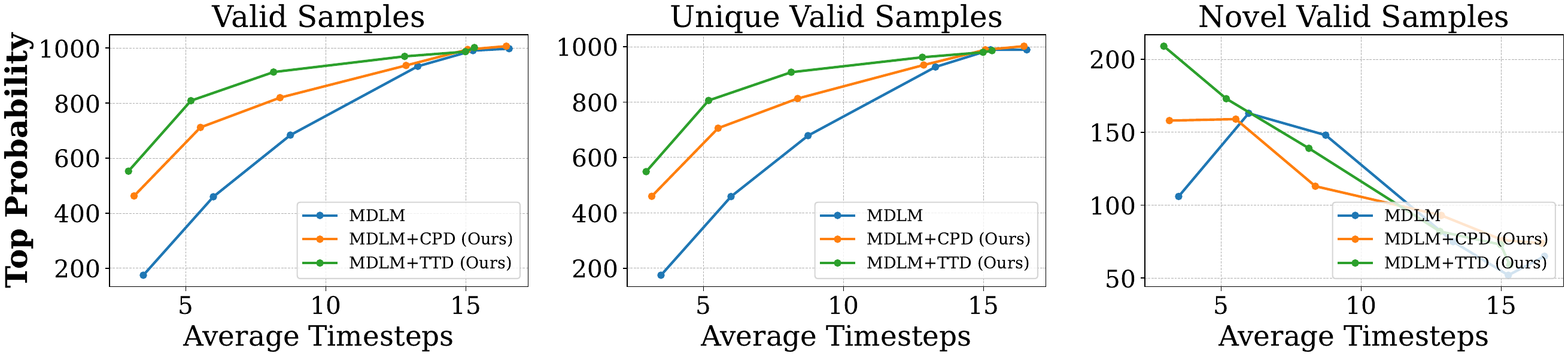}
  \caption{Molecule generation on QM9~\citep{blum2009qm9_1, rupp2012qm9_2} with EB-Sampler \citep{benhamu2025ebsampler}. ``Average Timesteps'' indicates the average number of timesteps taken by the sampler to generate samples.}
  \label{fig:QM9-EB}
\end{figure}

In~\Cref{fig:QM9-sampling-strategies} where we generate molecules with different ordering strategies, we observe that the gap between TTD and CPD is most clearly pronounced in left-to-right order generation. This agrees with our theory that TTD outperforms CPD due to its ability to model the dependence between neighboring tokens.
Orthogonal to the ordering strategy, we also adapt EB-Sampler~\citep{benhamu2025ebsampler}, which adaptively determines the number of unmasked~\emph{tokens} to validate whether our models can be paired with advanced samplers. As shown in~\Cref{fig:QM9-EB}, we observe that our methods outperform the baseline even with a different choice of sampler.

\begin{figure}
  \centering
  \includegraphics[width=1\linewidth]{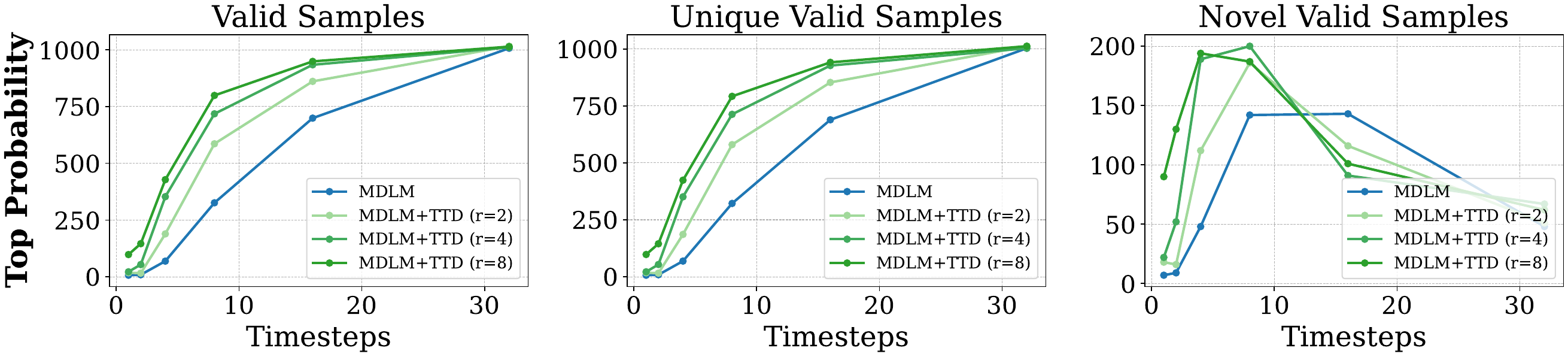}
  \caption{Rank size ablation on QM9 \citep{blum2009qm9_1, rupp2012qm9_2} }
  \label{fig:QM9-rank-scaling}
\end{figure}

From~\cref{fig:QM9-rank-scaling}, we can see that the performance of TTD monotonically increases with the number of ranks used, as expected.

\end{document}